\documentclass[acmtog]{acmart}
\usepackage{enumitem}
\usepackage{colortbl}
\usepackage{subfigure}
\usepackage{bm}
\usepackage{pifont}
\usepackage{multirow}
\usepackage{makecell}
\usepackage{xcolor} 
\definecolor{corporateBlue}{rgb}{0.1,0.2,1.0} 
\acmJournal{TOG}
\setcitestyle{square}

\newcommand{\Skip}[1]{}
\usepackage{xcolor}
\AtBeginDocument{%
  \providecommand\BibTeX{{%
    \normalfont B\kern-0.5em{\scshape i\kern-0.25em b}\kern-0.8em\TeX}}}

\begin{document}

\title{GS-PI: An Optimization-Decoupled Appearance Decomposition Approach for Generating PBR Gaussian Assets}
\author{Jieting Xu}
\affiliation{%
  \institution{State Key Laboratory of CAD\&CG, Zhejiang University}
  \city{Hangzhou}
  \country{China}}
\email{xujieting@zju.edu.cn}

\author{Rengan Xie}
\affiliation{%
  \institution{State Key Laboratory of CAD\&CG, Zhejiang University}
  \city{Hangzhou}
  \country{China}}
\email{rgxie@zju.edu.cn}

\author{Zijian Huang}
\affiliation{%
  \institution{State Key Laboratory of CAD\&CG, Zhejiang University}
  \city{Hangzhou}
  \country{China}}
\email{22421134@zju.edu.cn}

\author{Zehui Jin}
\affiliation{%
  \institution{State Key Laboratory of CAD\&CG, Zhejiang University}
  \city{Hangzhou}
  \country{China}}
\email{22521209@zju.edu.cn}

\author{Rui Wang}
\affiliation{%
  \institution{State Key Laboratory of CAD\&CG, Zhejiang University}
  \city{Hangzhou}
  \country{China}}
\email{ruiwang@zju.edu.cn}

\author{Yuchi Huo}
\authornotemark[2]
\authornote{\dag Corresponding author}
\affiliation{%
  \institution{State Key Laboratory of CAD\&CG, Zhejiang University}
  \city{Hangzhou}
  \country{China}}
\email{huo.yuchi.sc@gmail.com}


\renewcommand{\shortauthors}{Xu et al.}

\begin{abstract}
Gaussian Splatting (GS) excels at novel-view synthesis but encodes baked-in radiance, tightly entangling illumination with geometry and preventing seamless integration into physically based rendering (PBR) pipelines. Existing inverse-rendering methods attempt to disentangle materials via joint optimization, but often suffer from competing objectives that cause severe ambiguities and residual lighting artifacts. To overcome this, we present GS-PI, a novel optimization-decoupled framework that casts PBR material generation as a geometry-conditioned diffusion process on 3D point clouds. By operating directly in the 3D domain, our method inherently guarantees multi-view consistency, sidestepping the severe pixel correspondence issues that challenge 2D diffusion approaches. We introduce a multi-scale cross-view conditioning mechanism that integrates three complementary components: a global semantic prior, source-anchored photometric cues, and an absolute spatial learned view-direction conditioning signal. This design efficiently compresses complex multi-view evidence, mitigating cross-view projection misalignment and successfully preventing specular highlights from baking into intrinsic colors. By extracting a point cloud from a pre-trained Gaussian model, predicting PBR attributes via conditional diffusion, and distilling them back through differentiable rasterisation, we yield a fully relightable PBR-GS asset. GS-PI outperforms recent inverse-rendering baselines while replacing per-scene joint illumination/BRDF optimization with a learned diffusion pass followed by a short target-driven distillation, without requiring proxy meshes.
\end{abstract}



\begin{teaserfigure}
  \includegraphics[width=\textwidth]{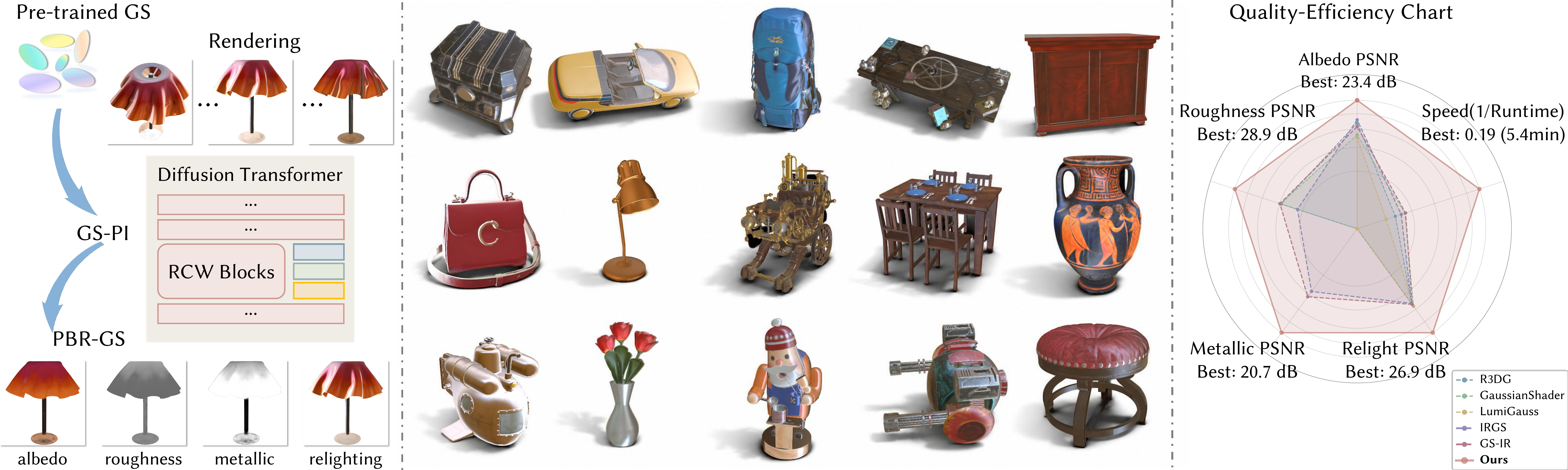}
  \caption{\textbf{GS-PI} is a generative pipeline for PBR material decomposition from pre-trained 3D Gaussians. Our method renders multi-view observations and predicts physically based material attributes through a diffusion transformer with \textbf{RCW} blocks, producing a relightable  \textbf{PBR-GS} with albedo, roughness, and metallic maps (Left). We present material decomposition results on diverse objects (middle) and compare reconstruction quality and efficiency with inverse-rendering baselines (right). Speed is defined as the reciprocal of runtime, and higher values indicate better performance.}
  \label{fig:teaser}
\end{teaserfigure}


\maketitle
\section{INTRODUCTION}\label{sec:intro}
Gaussian Splatting (GS)~\cite{kerbl_2023} has rapidly become a preferred representation for high-fidelity, real-time novel-view synthesis, and a growing number of feed-forward reconstructors~\cite{hong_2024,charatan_2024,tang_2024} can now produce Gaussian assets from sparse images in seconds. Despite this progress, the resulting primitives encode only view-dependent radiance---typically via spherical harmonics---tightly entangling illumination with geometry. Consequently, these assets cannot be relit, edited, or directly integrated into physically based rendering (PBR) pipelines. Closing this gap requires assigning physically meaningful material attributes (albedo, roughness, metallic) to each Gaussian primitive, a task we refer to as \emph{PBR material decomposition on Gaussians}.

Existing inverse-rendering approaches on Gaussians~\cite{liang_2024_GSIR,gao_2024,gu_2024,kaleta2025lumigauss,jiang2024gaussianshader} formulate material estimation as a joint optimization of illumination and BRDF parameters. To stabilize this under-constrained problem, they typically rely on hand-crafted regularisations tailored to specific object categories or material types, producing complex multi-objective optimization landscapes in which competing objectives interfere with the core inverse-rendering signal. As shown in Fig.~\ref{fig:baseline}, these conflicts introduce ambiguities that degrade material fidelity and leave lighting baked into the decomposed attributes. Integrating a full PBR rendering loop also makes such pipelines computationally expensive and exacerbates the convergence conflicts. There is therefore a clear need for a framework that decouples material generation from restrictive per-scene optimization.

Recently, learned diffusion models have shown strong results for 2D image generation, outperforming traditional optimization-based inverse rendering in fidelity and robustness. However, 2D generative priors cannot be directly transferred to 3D objects, as they lack spatial awareness and struggle to maintain multi-view consistency. Motivated by this, we extend the diffusion-prior paradigm to 3DGS material decomposition. Given a reconstructed Gaussian scene, our method decomposes the baked appearance into physically meaningful PBR attributes. Because material generation is formulated within a 3D point-cloud diffusion latent space, multi-view consistency is maintained by construction. To the best of our knowledge, this is the first work to adapt 3D diffusion for PBR material disentanglement on Gaussian primitives.

While 3D diffusion guarantees structural consistency, the generative process must still be grounded in the original photometric observations to accurately separate view-dependent specularities from intrinsic albedo. However, injecting multi-view image conditioning into point cloud diffusion models presents a critical bottleneck. Naively aggregating dense multi-view images causes feature explosion and exacerbates spatial blurring due to the inherent sub-pixel depth noise of reconstructed primitives. Conversely, relying on a single view discards the cross-view reflection cues essential for disentanglement, inevitably baking specular highlights into the albedo. We resolve this paradox by introducing a \textbf{multi-scale cross-view conditioning mechanism} that extracts features across three inseparable perceptual dimensions. First, a macroscopic global semantic prior \cite{oquab_2023} provides a foundational physical understanding to prevent the misinterpretation of specular highlights. Second, a microscopic deterministic routing strategy explicitly binds the optimal RGB observation to its source view, rigorously preserving essential micro-details without multi-view blurring. Finally, an explicit learned view-direction conditioning signal serves as an absolute directional anchor, establishing the critical mapping between viewing angles and specular peaks to avoid baking highlights into the intrinsic material.

Recall that joint inverse rendering inherently entangles shape updates with lighting estimation, leading to the aforementioned optimization ambiguities. To circumvent this, we propose a pipeline that explicitly decouples geometry from appearance. We first extract a stable geometric structure---represented as a point cloud---from an input Gaussian model, freezing spatial properties to isolate the material generation task. After predicting PBR attributes via conditional diffusion, we must map these point-based predictions back to the pre-trained Gaussian model. Thus, we distill these generated materials onto the original primitives through differentiable rasterization, yielding a fully relightable PBR-GS asset.

In summary, we make the following contributions:
\begin{enumerate}[topsep=0pt, itemsep=0pt, parsep=0pt]
    \item We present the first point-cloud-diffusion-based optimization-decoupled pipeline that converts an image-reconstructed Gaussian model into a relightable PBR asset, decoupling material generation from time-consuming per-scene inverse-rendering optimization over illumination and BRDF; the only per-scene step is a short, target-driven distillation that binds PBR maps onto primitives.
    \item We perform material disentanglement directly in a point-cloud diffusion latent space, maintaining multi-view consistency by construction and sidestepping the pixel correspondence issues that affect 2D diffusion methods.
    \item We propose a multi-scale cross-view conditioning mechanism that unites a macroscopic global semantic prior, microscopic source-anchored photometric cues, and a learned view-direction conditioning signal. This synergistic design effectively resolves the multi-view feature explosion and cross-view cue omission dilemmas on noisy geometries.
\end{enumerate}

\section{RELATED WORK}
\label{sec:related}

\subsection{Gaussian Splatting and Inverse Rendering}
Gaussian Splatting (GS)~\cite{kerbl_2023,huang_2024_2dgs} models scenes using unstructured primitives rendered via tile-based rasterization, achieving state-of-the-art novel-view synthesis at real-time frame rates. While feed-forward reconstructors~\cite{hong_2024,charatan_2024,tang_2024} can rapidly generate these models from sparse images, the resulting primitives encode view-dependent radiance. Retrofitting them with physically based materials is typically approached via inverse rendering~\cite{liang_2024_GSIR,gao_2024,gu_2024,kaleta2025lumigauss,jiang2024gaussianshader}. These methods formulate material estimation as a highly non-convex joint optimization problem with different objectives: GS-IR~\cite{liang_2024_GSIR} and IRGS~\cite{gu_2024} jointly optimize environmental lighting, visibility, and BRDF parameters, while GaussianShader~\cite{jiang2024gaussianshader} introduces specialized normal and shading regularisations for reflective surfaces. More recently, deferred-shading formulations such as 3DGS-DR~\cite{ye20243dgsdr} and DeferredGS~\cite{wang2024deferredgs} defer the BRDF evaluation to screen space to obtain sharper reflections and editable materials. To stabilize this under-constrained problem, such pipelines inject hand-crafted regularisations tied to specific object categories or heuristic material priors (e.g., spatial smoothness or monochromatic metallic constraints). These competing objectives frequently interfere with the core inverse-rendering signal, producing ambiguities that trap the optimizer in local minima and leaving specular highlights baked into the albedo, as visible in our baseline comparisons. In contrast, GS-PI avoids per-scene inverse-rendering optimization over illumination and BRDF.

\subsection{Point Cloud Diffusion Models}
Point-NeRF \cite{xu_2022} and Neural Point Cloud Diffusion (NPCD) \cite{schroppel_2024} established the viability of hybrid point-feature representations for denoising diffusion \cite{ho_2020,song_2021}. Concurrently, PointInfinity \cite{huang_2024_pointinfinity} introduced a resolution-invariant, two-stream Read--Compute--Write (RCW) transformer that elegantly decouples fixed-size latent tokens from variable-size point sets. We adopt this robust backbone but diverge significantly in our problem formulation. Existing point-cloud diffusion models jointly generate geometry and appearance, which is redundant and computationally prohibitive for dense PBR generation. Instead, GS-PI freezes the coordinate space to exploit the known geometry prior, conducting diffusion exclusively within the PBR feature space. Furthermore, scaling these architectures to handle dense multi-view inputs for material generation leads to a severe dimensionality explosion. To resolve this, we extend the two-stream backbone with our multi-scale cross-view conditioning mechanism, incorporating source-anchored photometric routing and a learned view-direction conditioning signal. This fundamentally adapts the architecture to disentangle physical materials robustly.

\subsection{Material Generation via 2D Diffusion Priors}
Recent works leverage 2D diffusion priors~\cite{rombach_2022} for material synthesis. One paradigm generates mesh PBR textures via iterative UV-space projection (e.g., TEXTure~\cite{richardson_2023}, Fantasia3D~\cite{chen_2023}, Paint3D~\cite{zeng_2024}, MatAtlas~\cite{ceylan2024matatlas}, DreamMat~\cite{zhang2024dreammat}). However, lacking UV parameterizations, unstructured Gaussian primitives are incompatible with these pipelines and suffer severe projection artifacts. A second paradigm guides inverse-rendering optimization via 2D priors for intrinsic decomposition (e.g., IntrinsicAnything~\cite{chen2024intrinsicanything}, MaterialAnything~\cite{huang2025material}). While relaxing the need for illumination ground truth, they remain slow due to per-scene optimization and lack 3D spatial awareness. In overlapping regions, inconsistent 2D score distillations cause conflicting gradients and blurred material estimates. Conversely, GS-PI performs material decomposition in a 3D point-cloud diffusion latent space. This optimization-decoupled 3D diffusion achieves spatially consistent predictions and significantly faster generation.

\subsection{Image-Conditioned 3D Generation}
Recent image-conditioned 3D generative models—including score distillation \cite{poole_2023,lin_2023}, novel-view diffusion \cite{liu_2023,liu_2024_syncdreamer,long_2024}, and feed-forward reconstructors \cite{hong_2024,tang_2024,charatan_2024}—excel at generating geometry and view-dependent RGB but ignore the PBR material in Gaussian splatting. GS-PI fills this gap by retrofitting these assets with PBR properties. To anchor our synthesis, we utilize DINOv2 \cite{oquab_2023} as a global semantic prior. Its robust dense-feature semantics provide the network with essential object-level physical understanding, serving as the foundational building block for our multi-scale cross-view conditioning mechanism.
\begin{figure*}[t]
  \centering
  \includegraphics[width=\linewidth]{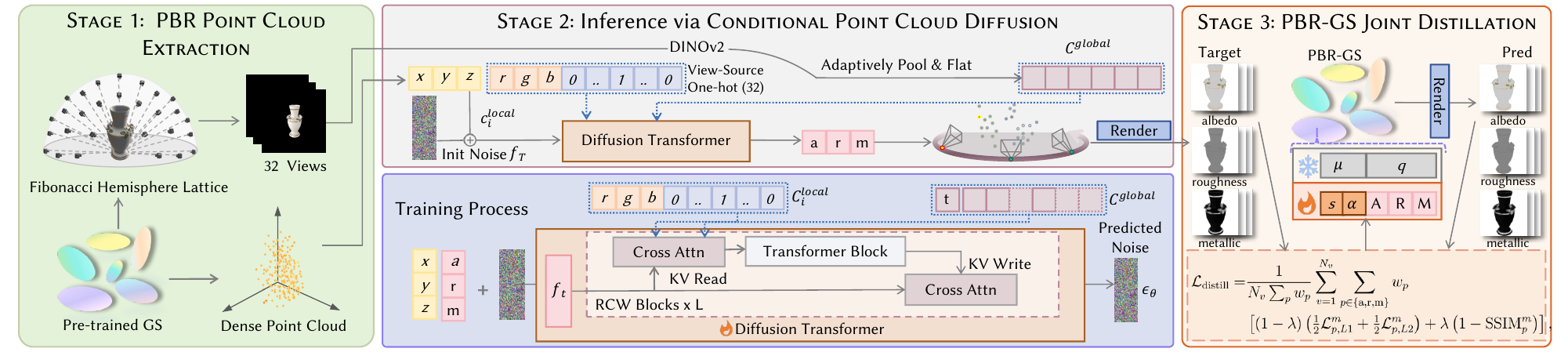}
   \caption{\textbf{Overview of the GS-PI Framework.} Our pipeline converts image-reconstructed Gaussians into relightable PBR assets through three sequential stages: \textbf{Stage I: Geometric Extraction}, where a consistent point-cloud substrate is distilled from the pre-trained Gaussian scene. \textbf{Stage II: Conditional Feature Diffusion}, where a diffusion transformer predicts PBR attributes on frozen geometry. This process is guided by a multi-scale conditioning triad—integrating global semantics, source-anchored photometric cues, and a view-direction conditioning signal —to rigorously disentangle specular highlights. \textbf{Stage III: Joint PBR Distillation}, which maps the inferred attributes back to the original Gaussians via masked differentiable rasterization and adaptive geometric relaxation. This optimization-decoupled paradigm achieves high-fidelity material decomposition while bypassing time-consuming per-scene inverse-rendering optimization over illumination and BRDF.}\label{fig:pipeline}
\end{figure*}

\section{METHOD}\label{sec:method}
We aim to transform image-reconstructed Gaussian models into fully relightable PBR assets without per-scene inverse-rendering optimization. To overcome dimensionality explosion and multi-view conflicts, we decouple geometry and appearance via a conditional point-cloud diffusion transformer model.

GS-PI comprises three stages (Fig.~\ref{fig:pipeline}): Section~\ref{sec:stage1} extracts a dense point cloud from pre-trained Gaussians. Section~\ref{sec:stage2} introduces our core conditional diffusion model, which uses multi-scale cross-view conditioning to compress dense multi-view photometric cues into a 35-dimensional per-point condition (3-dim RGB + 32-dim view-source one-hot). Finally, Section~\ref{sec:stage3} distills the inferred PBR attributes back to the original Gaussians via differentiable rasterization.

\subsection{Problem formulation.}
Given multi-view images $\{\mathbf{I}_v\}_{v=1}^{V}$ (where $V{=}32$ in our dense setting) of an object and its reconstructed Gaussian model $\mathcal{G}=\{\boldsymbol{\mu}_i,\mathbf{q}_i,\mathbf{s}_i,\alpha_i\}_{i=1}^{N_g}$, where $N_g$ is the number of primitives, our ultimate goal is to assign PBR attributes to these $N_g$ Gaussians. To bridge the gap between 3D generative priors and Gaussian primitives, we formulate our core generative process on an extracted point cloud of size $N$. Therefore, our direct prediction target is the per-point PBR attributes $\mathbf{F}\in\mathbb{R}^{5\times N}$. Each column $\mathbf{f}_i=[\mathbf{a}_i,\,r_i,\,m_i]^{\top}$ concatenates an albedo $\mathbf{a}_i\in[0,1]^{3}$, a roughness $r_i\in[0,1]$, and a metallic coefficient $m_i\in[0,1]$. The point coordinates $\mathbf{X}\in\mathbb{R}^{3\times N}$ remain fixed. Generating PBR attributes natively on this scale presents three distinct challenges: \textit{(i)} geometry must be exploited as an exact prior rather than co-generated; \textit{(ii)} PBR features are heterogeneous and strictly bounded; and \textit{(iii)} high-dimensional multi-view photometric observations must be efficiently compressed and spatial ambiguities fundamentally resolved to form a unified condition.

\subsection{Point Cloud Extraction}\label{sec:stage1}
While our pipeline is fundamentally agnostic to the underlying Gaussian representation and can be applied to any standard 3DGS variant, Stage I extracts a dense point cloud from a pre-trained 2D Gaussian Splatting (2DGS) model. We preferentially use 2DGS as its surfel-like primitives yield cleaner, surface-aligned depth maps with fewer internal floaters compared to volumetric 3DGS. For each source view, we unproject valid foreground pixels into world space based on rendered median depth and opacity masks. Each unprojected point directly captures the photometric RGB observation from its originating pixel. To suppress geometric noise and floaters, we apply global statistical outlier removal, cross-view mask consistency checks, and depth consensus refinement. The comprehensive details and hyperparameters of this geometric filtering pipeline are deferred to the Supplementary Material. Consequently, we obtain a dense point cloud coupled with direct RGB observations, establishing the foundation for Stage II.

\subsection{Conditional Point Cloud Diffusion Transformer}\label{sec:stage2}

Given the high density of the extracted point cloud, standard self-attention mechanisms with $\mathcal{O}(N^2)$ complexity are computationally prohibitive. Therefore, we base our diffusion backbone on the PointInfinity architecture \cite{huang_2024_pointinfinity}, which features a two-stream transformer design and achieves $\mathcal{O}(N)$ complexity by maintaining a compact set of global latent tokens $\mathbf{z}$. Specifically, the network alternates between reading features from the dense point cloud into the latent stream, computing global interactions among latents via dense self-attention, and writing the updated global context back to the individual points.

We adopt this architecture not only for its linear scalability but also because its latent stream naturally serves as a global information bottleneck. This elegantly pairs with our material-aware conditioning, facilitating long-range physical consistency (e.g., uniform metallic properties across a continuous surface) while avoiding localised noise. Despite these structural advantages for global coherence, the original framework was designed primarily for geometric generation. It inherently lacks the capacity to ingest complex photometric cues, or resolve the view-dependent ambiguities crucial for inverse rendering. To overcome these limitations and effectively tackle physical material disentanglement, we fundamentally re-architect its state space and conditioning mechanisms.

\subsubsection{Geometry-Conditioned, Feature-Only Diffusion}
Existing point-cloud diffusion models \cite{schroppel_2024} jointly diffuse coordinates and features, forcing the network to waste capacity reconstructing already-known geometry. We explicitly freeze the spatial coordinates $\mathbf{X}$: the forward process is applied exclusively to the PBR feature tensor:
\begin{equation}
\mathbf{f}_t = \sqrt{\bar{\alpha}_t}\,\mathbf{f}_0 + \sqrt{1-\bar{\alpha}_t}\,\boldsymbol{\epsilon},\quad \boldsymbol{\epsilon}\sim\mathcal{N}(\mathbf{0},\mathbf{I}),
\label{eq:feature_forward}
\end{equation}
and the denoiser $\boldsymbol{\epsilon}_\theta$ predicts only the 5-channel feature noise $\hat{\boldsymbol{\epsilon}}_{\mathrm{feat}}$. Freezing geometry reduces the target dimensionality and eliminates the need for geometric regularisation.
To ensure training stability and physically valid outputs, the frozen coordinates $\mathbf{X}$ and the bounded PBR features are heterogeneously normalised before entering the diffusion process (details are provided in the Supplementary Material).

\subsubsection{Multi-Scale Cross-View Conditioning Mechanism}
\label{sec:stage2_cond}

To condition the generative process on $V=32$ reference views without triggering feature explosion or multi-view blurring, we propose a multi-scale cross-view mechanism. This triad seamlessly integrates global, local, and spatial light cues into an ultra-compact per-point representation.

\paragraph{Global Semantic Prior.} 
Before inferring micro-surface details, the network must establish a foundational physical cognition (e.g., distinguishing "rusted metal" from "smooth plastic"). We leverage offline-precomputed features extracted by a frozen DINOv2 \cite{oquab_2023} encoder. To ensure computational tractability without sacrificing macroscopic context, these features are adaptively pooled into a $5\times5$ spatial grid per view. During training, we introduce a view-dropout strategy by stochastically subsampling 16 out of 32 views, which inherently regularises the network against view-overfitting. The retained tokens are flattened to form $\mathbf{C}^{\mathrm{global}}$ and linearly projected into the latent stream $\mathbf{z}$ before the RCW stack. This macro-prior anchors the baseline distribution for roughness and metallic attributes.

\paragraph{Microscopic Source-Anchored Photometric Cues (Top-RGB)}

Fusing 32 views per point inherently mixes high-fidelity observations with blurred or occluded pixels, destroying micro-details. To address this and explicitly clarify how microscopic features are collected, we implement a source-anchored deterministic routing strategy. Rather than dynamically computing view priorities based on orthogonality, we deterministically extract the exact RGB observation from the point's originating view. Each 3D point $\mathbf{x}_i$ is unprojected from a specific source view $v^\star$ during Stage I. We directly collect its microscopic photometric feature by sampling the RGB image at this exact projected location: $\mathbf{I}_{v^\star}(\pi_{v^\star}(\mathbf{x}_i))$, where $\pi_{v^\star}(\cdot)$ denotes the perspective projection operator mapping a 3D coordinate to the 2D image plane of view $v^\star$.

Because reconstructed primitives inherently contain sub-pixel depth noise, this explicit hard-binding guarantees that the geometric coordinate and its colour observation suffer from the exact same projection distortions, effectively neutralizing cross-view misalignment. One might intuit that aggregating more observations (e.g., fusing the top 6 visible views) would provide a richer sampling of the specular lobe. However, our ablation studies reveal that such naive multi-view aggregation introduces spatial variance that catastrophically blurs high-frequency material details. Thus, our aggressive single-view routing acts as a deliberate low-pass filter for geometric noise. To compensate for the omitted view-dependent photometric cues, the model relies on the macroscopic semantic priors to hallucinate plausible high-frequency specular distributions based on the recognized material semantics.

\paragraph{Learned View-Direction Conditioning Signal: Implicit Angle Awareness via View-Source Indicator.}
Relying solely on a single source-anchored RGB observation introduces a fatal physical flaw: the loss of cross-view reflection cues makes it impossible to differentiate an intrinsic white albedo from a baked-in specular highlight. We resolve this by appending a 32-dimensional one-hot view-source indicator $\mathbf{o}_i$ to the sampled RGB feature. Because the point stream retains the absolute 3D coordinates $\mathbf{X}$, this discrete 32-dim vector acts as a learned view-direction conditioning signal; it is not a physical estimate of the scene illumination, but a discrete angular index that lets the network associate viewing directions with the specular response it must explain away. It explicitly signals to the network the exact observation angle responsible for the sampled colour.
\begin{equation}
\mathbf{c}_i^{\mathrm{local}} = \bigl[\,\mathbf{I}_{v^\star}(\pi_{v^\star}(\mathbf{x}_i))\,\|\,\mathbf{o}_i\,\bigr] \in \mathbb{R}^{3+32}.
\label{eq:per_point_cond}
\end{equation}

Importantly, this canonical 32-view set is an internal rendering protocol rather than a constraint on the input capture. Given a reconstructed Gaussian asset, we deterministically rasterize virtual observations on the fibonacci hemisphere lattice and use the view-source one-hot vector only as an angular index on this internal lattice. Thus, our method remains compatible with any reconstruction pipeline that produces an initial Gaussian asset, including sparse-view 3DGS reconstruction methods.

Since these camera poses remain strictly invariant across all scenes, each one-hot index maps uniquely to an absolute physical viewing direction. This canonicalization enables the network to learn how canonical viewing directions correlate with view-dependent reflectance patterns. As shown in our ablation (Table ~\ref{tab:dataset256_ablation_material_metrics}, A2), removing this discrete spatial anchor degrades the network's ability to decouple illumination from albedo (PSNR 34.66 → 29.79), confirming that explicit angular conditioning is indispensable for specular disentanglement.

By synergising the global DINOv2 prior, the 3-dim Top-RGB, and the 32-dim view-direction conditioning signal, we elegantly compress 128 dimensions of multi-view redundancy into a highly effective \textbf{35-dimensional} local condition $\mathbf{c}_i^{\mathrm{local}}$. This triad empowers the RCW network to implicitly map viewing angles to specular peaks, cleanly disentangling highlights from the albedo.

We optimize the denoiser via an $\epsilon$-prediction DDPM objective \cite{ho_2020}. To stabilise convergence across the heterogeneous PBR feature space, we employ a cosine noise schedule and apply Min-SNR-$\gamma$ loss reweighting, which prioritises learning high-frequency material details over redundant low-frequency signals.

\subsubsection{Training vs. Inference Dynamics.}
Our diffusion framework exhibits critical differences between training and inference to ensure robust generalisation and memory efficiency. During training, the model operates in a single-step noise-prediction mode. To bridge the domain gap between ideal geometries and noisy reconstructed Gaussians, we do not train on perfect surfaces. Instead, we extract point clouds from ground-truth meshes and inject random spatial perturbations to simulate the inherent sub-pixel depth noise of Gaussian primitives. From this perturbed geometric corpus, we randomly sample a local subset of $N_{\mathrm{train}} = 512$ points per iteration to mitigate GPU memory constraints and prevent the network from overfitting to specific global shape priors. The PBR features of these sampled points $\mathbf{f}_0$ are corrupted to $\mathbf{f}_t$ via Eq.~(\ref{eq:feature_forward}), and the network optimizes the $\epsilon$-prediction objective. Crucially, we enforce the aforementioned view-dropout strategy (subsampling 16 out of 32 views) for the global DINOv2 prior, forcing the RCW latent stream $\mathbf{z}$ to learn view-invariant material representations. \textbf{During inference (sampling)}, the generation process starts from pure Gaussian noise $\mathbf{f}_T \sim \mathcal{N}(\mathbf{0}, \mathbf{I})$. In contrast to the sparse subset used in training, the frozen network processes the \emph{entire} dense point cloud ($N_{\mathrm{test}} = N$, typically hundreds of thousands of points extracted in Stage I) simultaneously. Thanks to the $\mathcal{O}(N)$ linear scalability of the RCW architecture, this full-resolution inference leverages the complete global context, ensuring seamless material consistency across the entire asset without memory bottlenecks. In this phase, view-dropout is disabled, and the network deterministically aggregates the full available contextual cues to synthesise high-fidelity PBR attributes.

\subsection{Distilling PBR Attributes onto Gaussian Primitives}
\label{sec:stage3}
While Stage~II successfully infers high-fidelity PBR attributes within the discrete point cloud, standard physical rendering engines require these parameters to be continuously bound to the spatial footprint of the original Gaussian representation. We address this via a joint differentiable distillation process.

\paragraph{Prior-Guided Parameter Initialisation and Adaptive Geometric Relaxation.}
To establish a robust starting state for distillation, we initialise the logit-space albedo, roughness, and metallic parameters of the 2D Gaussian model $\mathcal{G}$ via multi-view back-projection. Specifically, each Gaussian centre is projected into the $V$ canonical reference views, gathering valid foreground samples bilinearly from the diffusion-generated target maps. For the albedo channel, this initialization is harmonised with the base colour decoded from the 0-th order spherical harmonics (SH) coefficients of the original reconstruction, ensuring that low-frequency radiance is preserved. During the subsequent optimization, the geometric means $\boldsymbol{\mu}$ and rotations $\mathbf{q}$ are strictly frozen to maintain topological integrity. However, we allow the opacities $\alpha$ and 2D scales $\mathbf{s}$ to undergo a conservative fine-tuning. This \emph{adaptive geometric relaxation} permits the Gaussians to subtly adjust their footprints to better accommodate the newly inferred high-frequency material boundaries.

\paragraph{Masked Differentiable Rasterization.}
Let $\tilde{\mathbf{A}}_v$, $\tilde{\mathbf{R}}_v$, and $\tilde{\mathbf{M}}_v$ denote the staged per-view PBR targets generated by Stage II. Using a differentiable 2DGS rasteriser, we render the corresponding Gaussian-bound PBR maps $\hat{\mathbf{A}}_v$, $\hat{\mathbf{R}}_v$, and $\hat{\mathbf{M}}_v$. We optimize the material parameters by minimizing a masked photometric and structural objective across all views:

\begin{equation}
\resizebox{\columnwidth}{!}{$
\mathcal{L}_{\mathrm{distill}} = \frac{1}{N_v \sum_{p} w_p} \sum_{v=1}^{N_v} \sum_{p\in\{\mathrm{a},\mathrm{r},\mathrm{m}\}} w_p \left[ (1-\lambda)\left(\frac{1}{2}\mathcal{L}^{m}_{p,L1}+\frac{1}{2}\mathcal{L}^{m}_{p,L2}\right) + \lambda\left(1-\mathrm{SSIM}^{m}_p\right) \right]
$}
\end{equation}

where $\mathcal{L}^{m}_{p,L1}$, $\mathcal{L}^{m}_{p,L2}$, and $\mathrm{SSIM}^{m}_p$ represent the foreground-masked $L_1$, $L_2$, and structural similarity metrics, respectively. The property weights are empirically set to $(w_{\mathrm{a}}, w_{\mathrm{r}}, w_{\mathrm{m}}) = (1, 1, 3)$; upweighting the metallic channel mitigates the gradient sparsity inherent to near-binary metallic distributions. Crucially, the binary foreground masks $\mathbf{B}_v$ (derived from rendered silhouettes) are strictly applied to suppress background contamination and block edge-bleeding artifacts, ensuring that material gradients are concentrated exclusively on the valid object support. The final output $\mathcal{G}_{\mathrm{PBR}}$ is a fully disentangled, relightable asset ready for deployment in any downstream physically based rendering pipeline.
\section{Experiments}\label{sec:eval}

\subsection{Experimental Setup}\label{sec:eval_setup}
\subsubsection{Datasets.} 
\begin{figure}[t]
  \centering
  \includegraphics[width=\columnwidth]{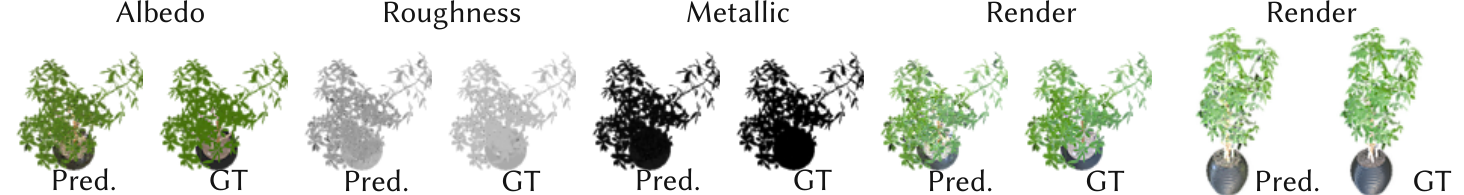}%
  \caption{\textbf{Example from NeRF-Synthetic dataset.} The recovered PBR maps yield consistent relighting under novel illumination.}  \label{fig:ficus}
\end{figure}

\begin{figure}[t]
  \centering
  \includegraphics[width=\columnwidth]{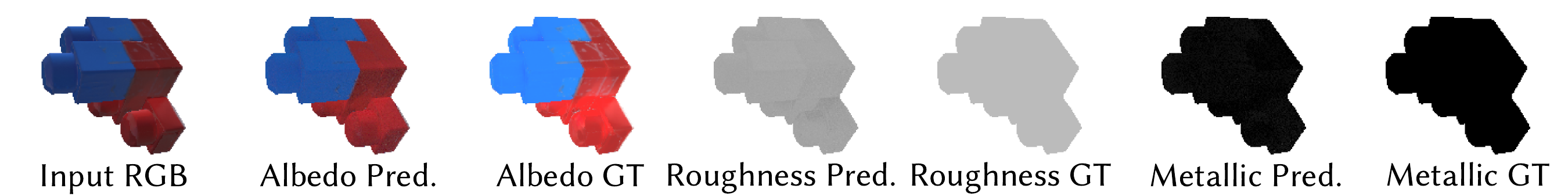}%
  \caption{\textbf{Qualitative results on real captures from the Stanford-ORB benchmark.} We recover plausible material on real-world scanned objects.}
  \label{fig:orb}
\end{figure}

We use synthetic PBR datasets (NeRF Synthetic (Fig.~\ref{fig:ficus}), game assets, Objaverse \cite{deitke2022objaverseuniverseannotated3d}) with 32-view renderings for quantitative benchmarking against ground-truth materials. 
We preliminary evaluate our method under real-wold data using the Google Scanned Objects (GSO)\cite{downs2022google}  and the Stanford-ORB benchmark~\cite{kuang2023stanfordorb} (Fig.~\ref{fig:orb}).

\subsubsection{Implementation Details.} During inference, Stage II processes a scene in ${\sim}18$ seconds (50 DDIM steps). Stage III then distills the PBR attributes via 500 optimization iterations in 4- 5 minutes. Detailed network architectures and hyperparameters are in the Supplementary Material.

\subsubsection{Data and PBR ground truth.} Our synthetic assets are rendered with a physically based split-sum shader, so albedo, roughness, and metallic maps serve as direct per-pixel material ground truth; relighting ground truth is obtained by re-rendering each asset with the same shader under held-out HDR environments. The 32 reference views used by both training and evaluation are deterministically rasterized on a fibonacci hemisphere lattice from the reconstructed Gaussian asset, and are an internal rendering protocol rather than a constraint on the original capture.

\subsubsection{Geometric filtering.} During Stage I point-cloud extraction, we remove floaters and geometric noise with three cascaded filters: (i) global statistical outlier removal, (ii) cross-view foreground-mask consistency checks, and (iii) depth-consensus refinement across neighboring views. The exact process are listed in the Supplementary Material.

\subsubsection{Baseline setup.} All inverse-rendering baselines are run with their official code and default configurations, and are relit under the same held-out HDR environments as GS-PI. 

\subsubsection{Baselines}\label{sec:eval_baselines}
\begin{figure}
  \centering
  \includegraphics[width=\columnwidth]{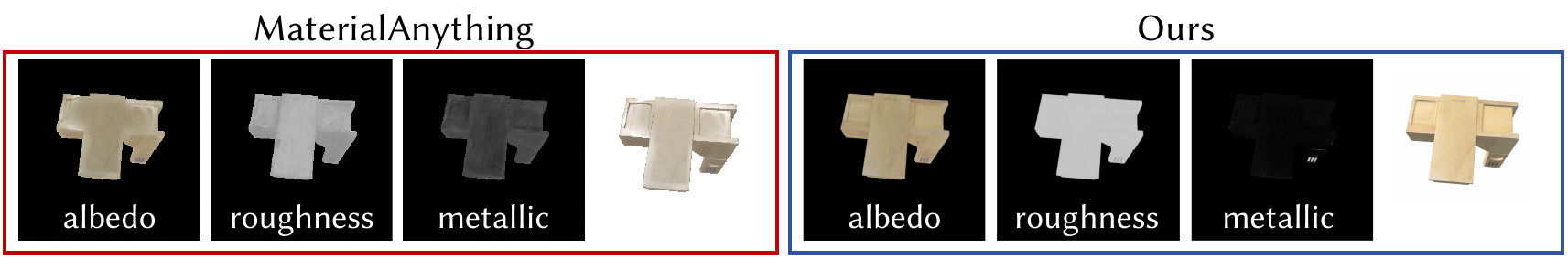} 
  \caption{\textbf{Qualitative comparison on the Google real-world dataset.} Compared to the baseline (MaterialAnything), our method reconstructs more physically plausible material maps, resulting in better rendering quality.} 
  \label{fig:baseline2}
\end{figure}

Although IntrinsicAnything targets native Gaussian PBR synthesis, we omit it as a baseline because its model checkpoints and multiview implementations remain closed-source. Consequently, in the absence of accessible generative models tailored for this task, we benchmark our approach against four state-of-the-art optimization-based inverse-rendering methods built upon 3D/2DGS (Fig.~\ref{fig:baseline}). However, to validate real-world generalization, we adapt the generative approach MaterialAnything for a comparative evaluation on the GSO dataset (Fig.~\ref{fig:baseline2}).

To isolate material quality from geometric noise, we introduce a \textbf{decoupled evaluation protocol}. During split-sum IBL relighting \cite{liang_2024_GSIR}, we render predicted materials using ground-truth normals. This prevents normal artifacts from confounding the evaluation of material disentanglement. This protocol is applied uniformly to all methods, including GS-PI: every entry in Table 1 uses the same ground-truth normals for split-sum IBL relighting, ensuring that the reported numbers reflect material disentanglement quality alone rather than differences in geometric reconstruction. 

\subsubsection{Visual comparison}
\begin{figure*}[ht!]
    \centering
    \resizebox{!}{0.93\textheight}{%
      \includegraphics{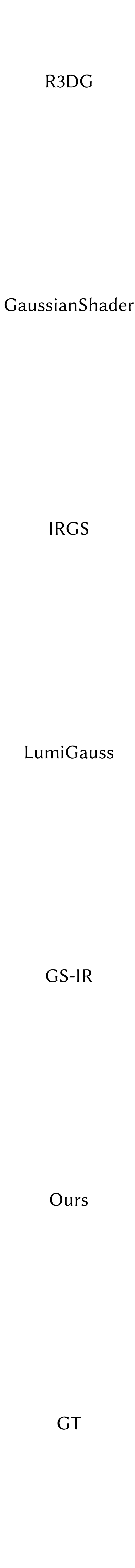}%
      \includegraphics{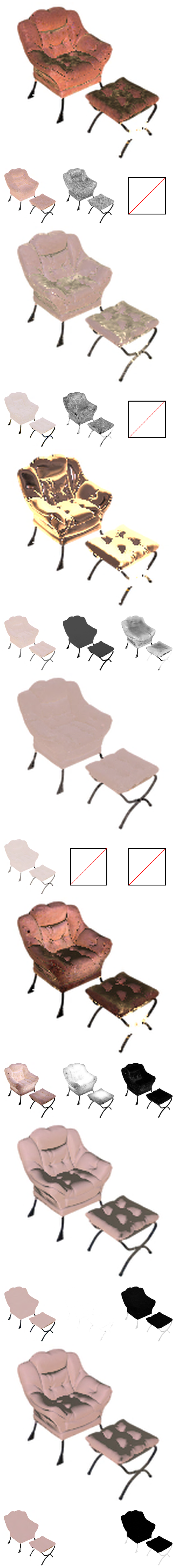}%
      \includegraphics{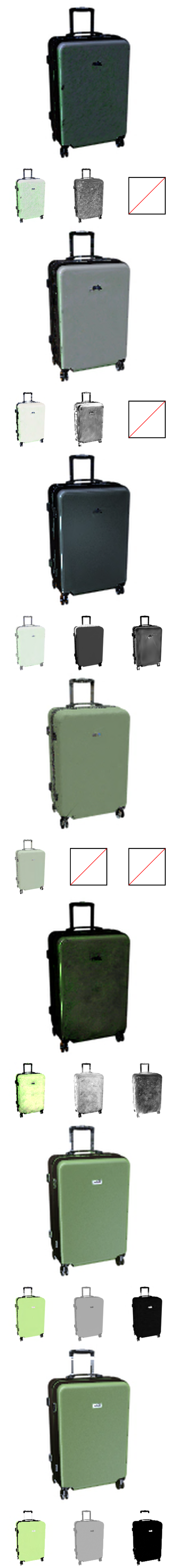}%
      \includegraphics{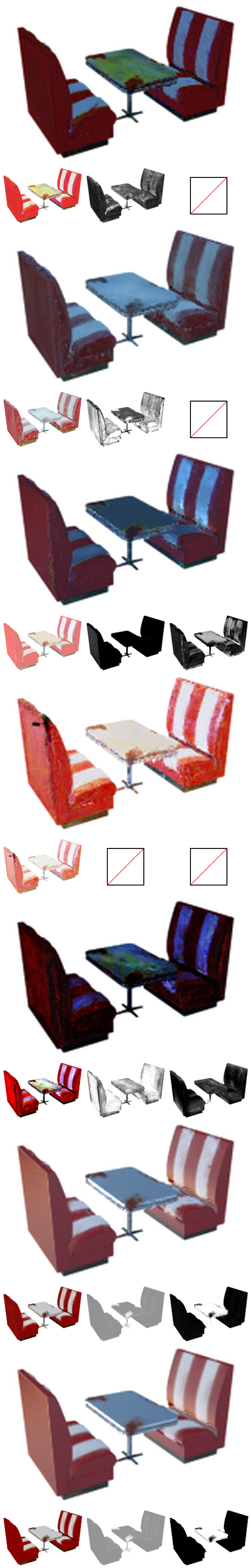}%
      \includegraphics{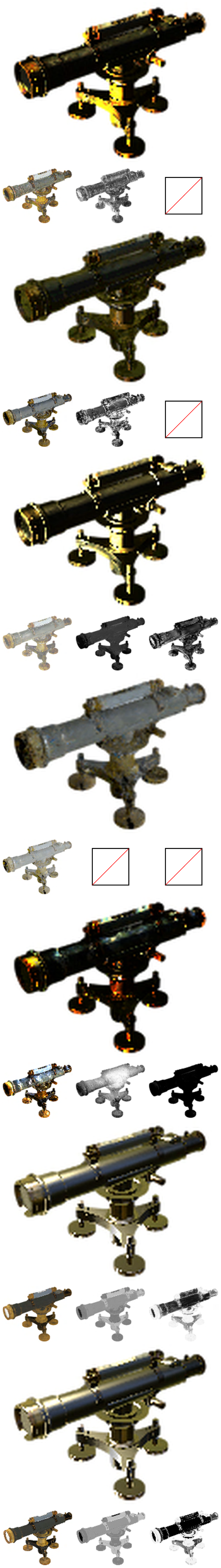}%
      \includegraphics{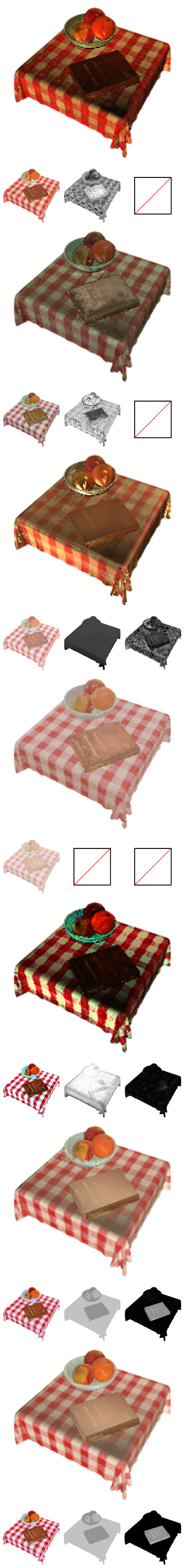}%
      \includegraphics{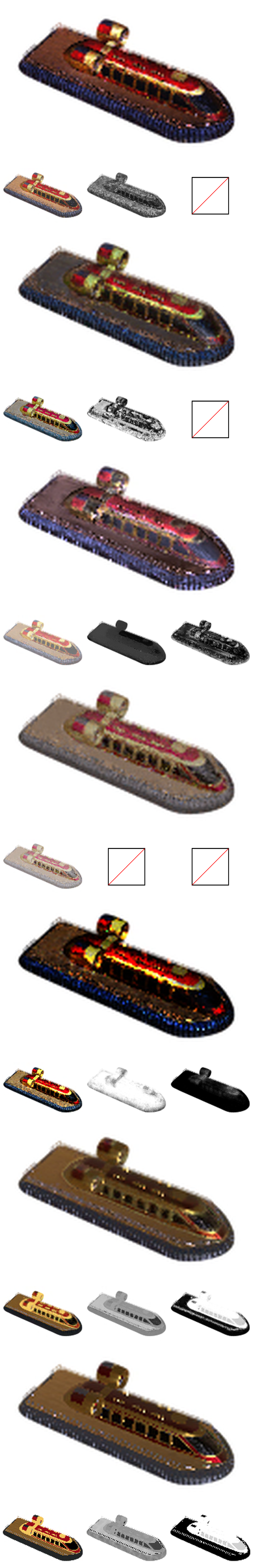}
    }
      \caption{\textbf{Qualitative Comparison.} We compare our GS-PI against representative inverse-rendering baselines including R3DG, GaussianShader, IRGS, LumiGauss, and GS-IR. For each method, the upper row displays the relighting result under novel illumination, while the lower row provides the decomposed material attributes: albedo (left), roughness (center), and metallic (right). Red crossed boxes indicate that the specific material attribute is not modeled by the corresponding baseline. Our method yields clean, physically plausible attributes that closely match the Ground Truth (GT), ensuring consistent relighting across diverse asset categories. We omit IntrinsicAnything as its checkpoint and related multiview code remain closed-source.}
    \Description{xxx}
    \label{fig:baseline}
\end{figure*}

Qualitatively, as illustrated in Fig.~\ref{fig:baseline}, existing inverse-rendering baselines consistently struggle to disentangle the inherent ambiguity between albedo and lighting. This limitation manifests as baked-in specular highlights erroneously projected onto their albedo predictions. By contrast, GS-PI leverages its structural diffusion prior and learned view-direction conditioning signal to explicitly separate highlights, yielding uniformly clean albedos and sharp, physically accurate metallic boundaries. This robust disentanglement allows our assets to generalize seamlessly to unseen environmental illumination, maintaining plausible specular flows. Our GS-PI is evaluated under the same protocol, i.e., the same ground-truth normals are used for relighting.
\begin{figure}[t]
  \centering
  \includegraphics[width=\columnwidth]{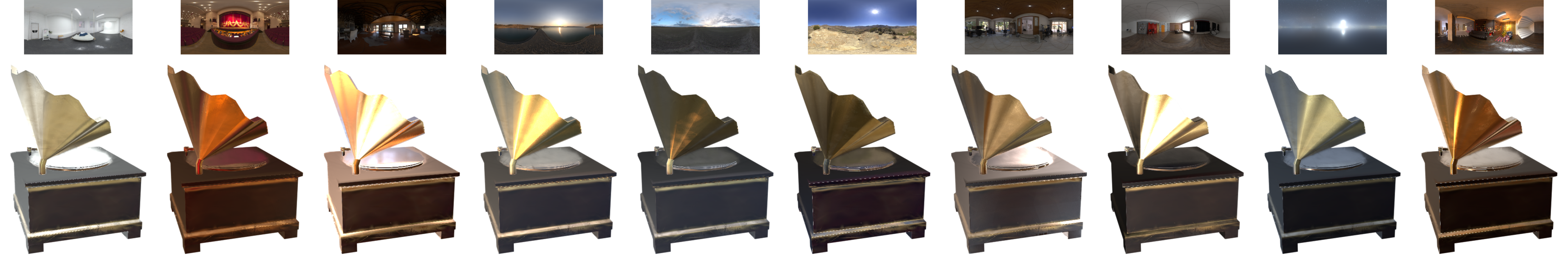}%
  \caption{\textbf{Relighting Results.} Our extracted PBR assets can be seamlessly relit under arbitrary environmental lighting. The figure displays a  gramophone rendered (below) with ten distinct HDR environment maps (top), demonstrating highly realistic material responses.}  \label{fig:relight}
\end{figure}
\begin{figure}[t]
  \centering
  \includegraphics[width=\columnwidth]{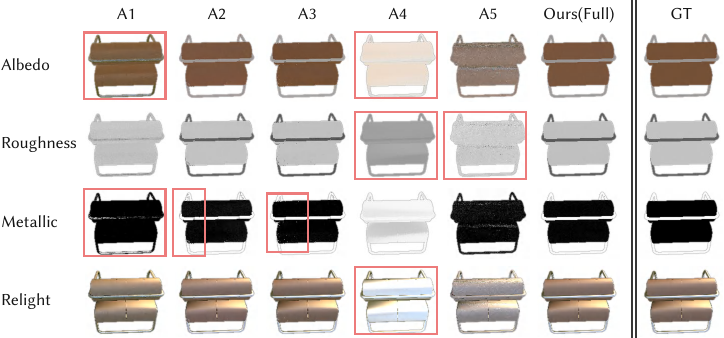} 
  \caption{\textbf{Qualitative ablation of Stage II.} Our full model ensures physically plausible disentanglement.}
  \label{fig:ablation_stage2}
\end{figure}
\begin{figure}
  \centering
  \includegraphics[width=\columnwidth]{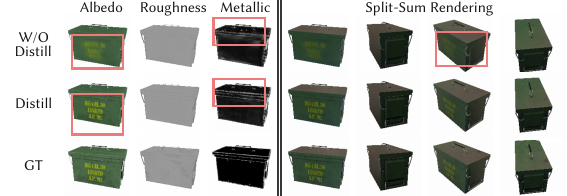} 
  \caption{\textbf{Qualitative ablation of Stage III distillation.} Bypassing distillation rigidly constrains primitives, leaving baked-in highlights on the albedo and structural artifacts in metallic/relighting maps. Our adaptive relaxation absorbs high-frequency details, preserving clean material boundaries.}  \label{fig:ablation_stage3}
\end{figure}
\begin{figure}[t]
  \centering
  \includegraphics[width=\columnwidth]{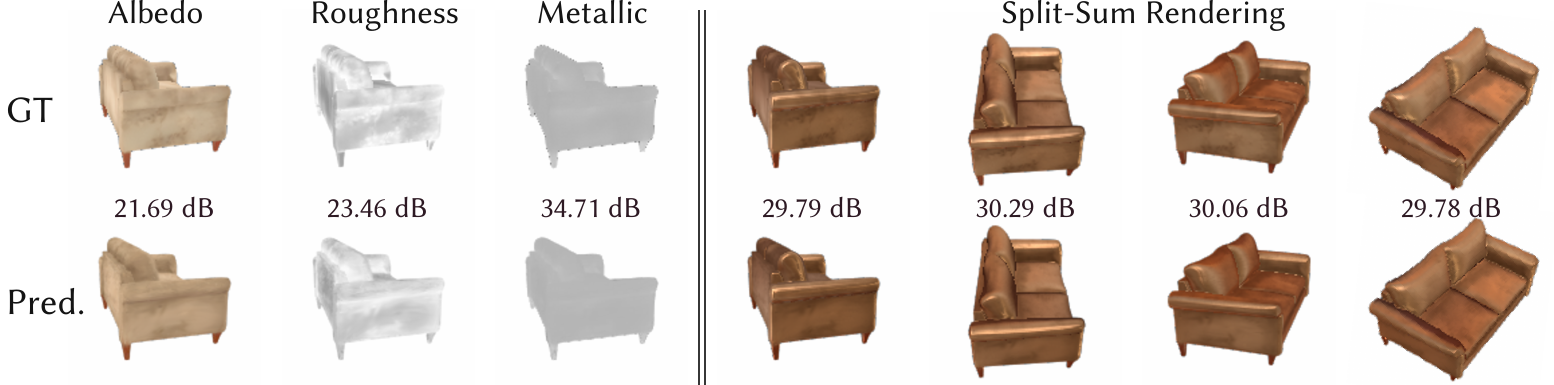} \\[-0.1cm]
  \includegraphics[width=\columnwidth]{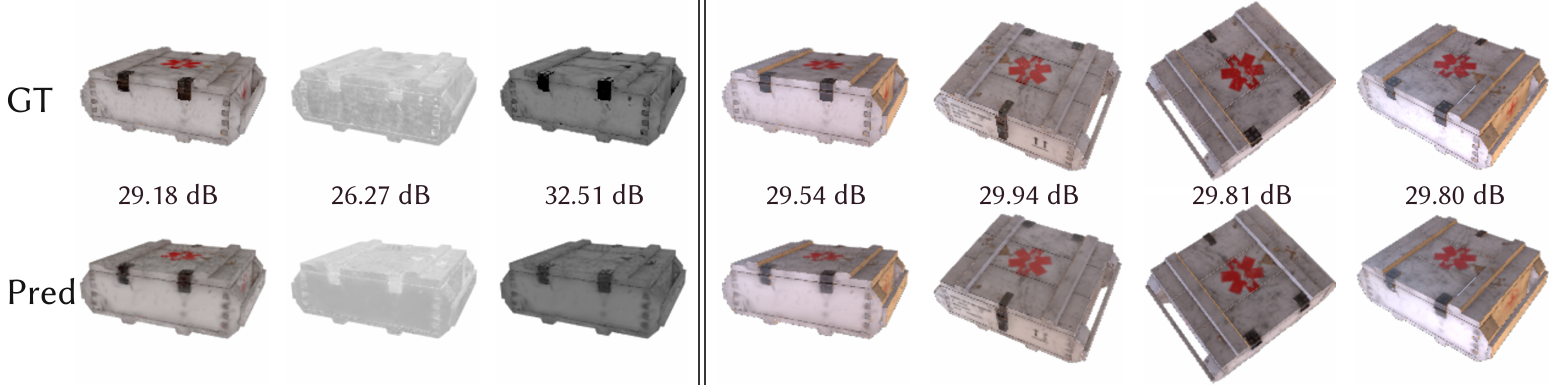}
  \caption{More results. For each 3D asset, the left panel displays the decomposed PBR material attributes. The right panel demonstrates the relighting capabilities across four viewpoints. The numerical values below each column denote PSNR, demonstrating the high-fidelity reconstruction of our method in both material disentanglement and novel-view relighting.}  
  \label{fig:relight_main}
\end{figure}
\begin{figure}[t]
  \centering
  \includegraphics[width=\columnwidth]{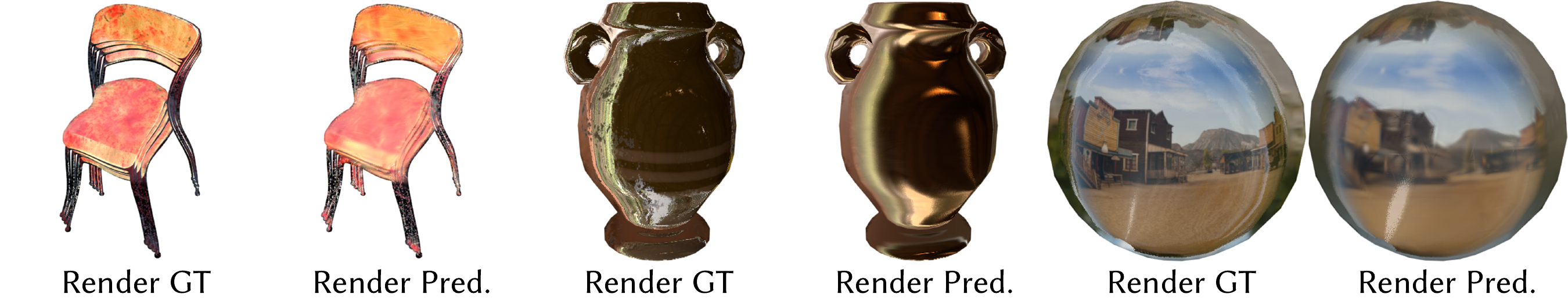}
  \caption{Failure cases. Each pair shows the ground-truth rendering (left) and our prediction (right). From left to right, the examples illustrate errors caused by incomplete reconstruction, over-smoothed high-frequency surface details, and inaccurate modeling of complex mirror-like reflections.}
  \label{fig:failure_case}
\end{figure}
\subsubsection{Quantitative Comparison}
To ensure a fair evaluation and rigorously isolate material accuracy from geometric artifacts, we report Peak Signal-to-Noise Ratio (PSNR), Structural Similarity Index (SSIM), and Learned Perceptual Image Patch Similarity (LPIPS) \cite{Zhang2018TheUE}. Metrics are evaluated on material maps, and novel-view relit images are computed exclusively within the ground-truth foreground mask. PSNR ~\cite{hore2010image} quantifies the absolute pixel-wise photometric fidelity, SSIM ~\cite{hore2010image} assesses the structural integrity of generated high-frequency details, and LPIPS provides a deep feature-based measure of perceptual similarity. Consequently, higher PSNR and SSIM, paired with lower LPIPS scores, definitively indicate superior material generation quality and physical accuracy. Fully coupled end-to-end quantitative evaluations (utilizing their self-predicted normals and their own shader) are in the Supplementary Material. 

As detailed in Table~\ref{tab:dataset256_baseline_material_metrics}, our method significantly outperforms all baselines by a large margin across every image-quality metric. Furthermore, our optimization-decoupled pipeline demonstrates a profound advantage in computational efficiency. Here the only per-scene step is a short, target-driven distillation rather than joint illumination/BRDF optimization. It requires only 5.4 minutes per scene, achieving a $6\times$ to $17\times$ speedup compared to optimization-based baselines (e.g., 34.4 minutes for GS-IR and 95.8 minutes for LumiGauss). These quantitative results robustly corroborate our visual findings, confirming that our diffusion-based approach not only effectively eliminates the specular leakage prevalent in traditional optimization paradigms but does so at a fraction of the computational cost.

\definecolor{BaselineBestCell}{RGB}{255,235,238}
\definecolor{BaselineSecondCell}{RGB}{255,249,196}
\newcolumntype{M}[1]{>{\centering\arraybackslash}m{#1}}

\begin{table*}[h]
    \centering
    \caption{Quantitative comparison on 1,000 images.  For Runtime, the values outside parentheses represent the material estimation time, while the values inside $(\cdot)$ denote the total end-to-end time including the initial 3D Gaussian reconstruction.}
    \label{tab:dataset256_baseline_material_metrics}
    \setlength{\tabcolsep}{0pt}
    \renewcommand{\arraystretch}{0.95}
    \small
    \begin{tabular*}{\textwidth}{@{\extracolsep{\fill}} l cccc cccc cccc c @{}}
        \toprule
        \multirow{2}{*}{Method} & \multicolumn{4}{c}{PSNR$\uparrow$} & \multicolumn{4}{c}{SSIM(\%)$\uparrow$} & \multicolumn{4}{c}{LPIPS(\%)$\downarrow$} & \multirow{2}{*}{\makecell{Runtime\\(min)}} \\
        \cmidrule(lr){2-5} \cmidrule(lr){6-9} \cmidrule(lr){10-13}
         & albedo & roughness & metallic & relight & albedo & roughness & metallic & relight & albedo & roughness & metallic & relight & \\
        \midrule
        
        R3DG 
        & 15.84 & 11.13 & -- & 14.00 
        & 57.08 & 27.42 & -- & 51.12 
        & 4.90 & 8.78 & -- & 5.03 
        & 47.5(55.3) \\
        
        GaussianShader 
        & 12.46 & 11.43 & -- & 14.97 
        & 53.30 & 29.32 & -- & 48.87 
        & 5.50 & 9.32 & -- & 5.84 
        & 27.4(35.4) \\
        
        LumiGauss 
        & 11.82 & -- & -- & 14.79 
        & 46.71 & -- & -- & \cellcolor{BaselineSecondCell}51.45 
        & 5.44 & -- & -- & \cellcolor{BaselineSecondCell}4.99 
        & 78.4(95.8) \\
        
        IRGS 
        & \cellcolor{BaselineSecondCell}16.80 & 6.87 & 7.53 & 13.88 
        & 63.95 & 31.85 & 6.69 & 49.62 
        & \cellcolor{BaselineSecondCell}4.59 & 10.11 & \cellcolor{BaselineSecondCell}12.46 & 5.30 
        & 36.5(41.6) \\
        
        GS-IR 
        & 14.79 & 11.51 & \cellcolor{BaselineSecondCell}8.85 & 14.26 
        & 48.81 & 39.46 & \cellcolor{BaselineSecondCell}12.75 & 49.19 
        & 6.30 & 8.84 & 13.60 & 5.68 
        & 29.8(34.4) \\
        
        MaterialAnything
        & 15.25 & \cellcolor{BaselineSecondCell}16.67 & 6.85 & \cellcolor{BaselineSecondCell}17.62
        & \cellcolor{BaselineSecondCell} 75.87 & \cellcolor{BaselineSecondCell}65.23 & 11.26 & 49.72 
        & 5.62 & \cellcolor{BaselineSecondCell} 5.65 & 12.65 & 5.52 
        & \cellcolor{BaselineSecondCell}22.4 \\
        
        \midrule
        \textbf{Ours} 
        & \cellcolor{BaselineBestCell}\textbf{23.38} & \cellcolor{BaselineBestCell}\textbf{28.86} & \cellcolor{BaselineBestCell}\textbf{20.68} & \cellcolor{BaselineBestCell}\textbf{26.94} 
        & \cellcolor{BaselineBestCell}\textbf{83.28} & \cellcolor{BaselineBestCell}\textbf{91.05} & \cellcolor{BaselineBestCell}\textbf{43.93} & \cellcolor{BaselineBestCell}\textbf{91.73} 
        & \cellcolor{BaselineBestCell}\textbf{2.90} & \cellcolor{BaselineBestCell}\textbf{1.69} & \cellcolor{BaselineBestCell}\textbf{8.62} & \cellcolor{BaselineBestCell}\textbf{1.39} 
        & \cellcolor{BaselineBestCell}\textbf{5.4(10.4)} \\
        \bottomrule
    \end{tabular*}
\end{table*}

\subsection{Ablation Study}\label{sec:eval_ablation}
To demonstrate the necessity of each architectural component, we conduct comprehensive ablation studies on the core design choices, summarizing the performance metrics in Table~\ref{tab:dataset256_ablation_material_metrics} and Table~\ref{tab:dataset256_distill_freeze_material_metrics_pbr_avg}. Conditioning Triad (A1, A2): As qualitatively shown in Fig.~\ref{fig:ablation_stage2}, omitting the global DINOv2 prior (A1) removes foundational semantic anchoring, causing unnatural albedo color shifts and erroneous metallic activations along object boundaries. Disabling the learned view-direction conditioning signal (A2) deprives the network of explicit view-directional awareness, hindering its ability to resolve albedo-lighting ambiguity and introducing salt-and-pepper noise in metallic maps. Both are vital for isolating specularities. Network Capacity \& Multi-View Injection (A3-A5): Naively scaling up the point sampling count (A3) or latent feature dimensions (A4) disrupts the spatial receptive field and induces optimization instability, leading to grain or mode collapse (yielding washed-out relighting in Fig.~\ref{fig:ablation_stage2}). Furthermore, directly injecting unaligned multi-view RGB observations (A5) overloads the conditioning space with conflicting projection cues due to inherent sub-pixel noise, causing chaotic noise rather than coherent materials. Stage III Distillation (A6-A9): As illustrated in Fig.~\ref{fig:ablation_stage3}, bypassing distillation (A6) leads to a significant loss of high-frequency information, yielding blurred text and "smudged" metallic boundaries. Strictly fixing the opacity (A7), scaling (A8), or both (A9) introduces structural rigidity. Our full adaptive relaxation strategy allows primitives to subtly adjust and "absorb" the high-frequency material details from the diffusion prior, ensuring perfect alignment with the underlying geometry.
\begin{table}[t]
    \centering
    \caption{Ablation study of the Stage II conditional diffusion model. We report average metrics for albedo, roughness, and metallic attributes predicted on the generated point cloud. All metrics are computed exclusively within the ground-truth alpha mask.}
    \label{tab:dataset256_ablation_material_metrics}
    \setlength{\tabcolsep}{0pt} 
    \renewcommand{\arraystretch}{0.95}
    \small
    \begin{tabular*}{\columnwidth}{@{\extracolsep{\fill}} l ccc @{}}
        \toprule
        Stage 2 Variant & PSNR$\uparrow$ & SSIM(\%)$\uparrow$ & LPIPS(\%)$\downarrow$ \\
        \midrule
        A1: w/o global semantic prior
        & 15.79 & 75.40 & 10.43 \\
        
        A2: w/o view-direction conditioning signal
        & \cellcolor{BaselineSecondCell}29.79 & 73.41 & 6.07 \\
        
        A3: $N_{pts}=1024$
        & 28.32 & \cellcolor{BaselineSecondCell}79.76 & \cellcolor{BaselineSecondCell}5.78 \\
        
        A4: $N_{pts}=1024$, $N_{lat}=384$ 
        & 8.33 & 50.65 & 10.05 \\
        
        A5: Direct multi-view injection
        & 13.31 & 33.65 & 16.07 \\
        
        \textbf{Ours (Full)} 
        & \cellcolor{BaselineBestCell}\textbf{34.66} & \cellcolor{BaselineBestCell}\textbf{80.46} & \cellcolor{BaselineBestCell}\textbf{3.95} \\
        \bottomrule
    \end{tabular*}
\end{table}
\begin{table}[t]
    \centering
    \caption{Ablation study of the Stage III distillation process, which maps the high-quality material priors generated in Stage II back onto the Gaussian representation. All metrics are evaluated between the distilled outputs and the alpha-masked ground-truth PBR maps.}
    \label{tab:dataset256_distill_freeze_material_metrics_pbr_avg}
    \setlength{\tabcolsep}{0pt} 
    \renewcommand{\arraystretch}{0.95}
    \small
    \begin{tabular*}{\columnwidth}{@{\extracolsep{\fill}} l ccc @{}}
        \toprule
        Stage 3 Variant & PSNR$\uparrow$ & SSIM(\%)$\uparrow$ & LPIPS(\%)$\downarrow$ \\
        \midrule
        A6: w/o distill 
        & 24.43 & 68.78 & 7.12 \\
        
        A7: Freeze opacity 
        & \cellcolor{BaselineSecondCell}26.28 & \cellcolor{BaselineSecondCell}78.21 & \cellcolor{BaselineSecondCell}5.19 \\
        
        A8: Freeze scaling 
        & 26.09 & 77.52 & 5.25 \\
        
        A9: Freeze opacity \& scaling 
        & 25.88 & 77.03 & 5.36 \\
        
        \textbf{Ours (Full)} 
        & \cellcolor{BaselineBestCell}\textbf{26.46} & \cellcolor{BaselineBestCell}\textbf{78.63} & \cellcolor{BaselineBestCell}\textbf{5.09} \\
        \bottomrule
    \end{tabular*}
\end{table}

\section{LIMITATIONS and CONCLUSIONS}\label{sec:conclusion} \subsection{Limitations} GS-PI has several limitations. First, although our learned view-direction conditioning is robust to moderate camera-layout changes, it still relies on a canonical lattice of $V=32$ virtual views, an internal rendering protocol rather than a capture constraint. GS-PI remains constrained by the completeness and geometry of the input Gaussian reconstruction, since Stage II uses frozen geometry and Stage III only conservatively relaxes opacity and scale. The left case in Fig.~\ref{fig:failure_case} illustrates this effect: incomplete reconstruction around slender structures provides insufficient point-cloud support, degrading local material decomposition and relighting. Thus, material quality remains bounded by the completeness and geometry of the reconstructed Gaussian asset. Second, translucent, refractive, subsurface, and near-perfect mirror materials fall outside our current normal/BRDF representation, causing blurred reflections or ambiguous material decomposition. Third, without tangent-space normal residuals, fine bumps, micro-normal variations, and tiny scratches can be over-smoothed (Fig.~\ref{fig:failure_case}). Future work could incorporate micro-normal prediction, flexible geometry refinement, and more complex light transport such as spatially varying refraction and subsurface scattering.

\subsection{Conclusion} GS-PI, the first framework to formulate Gaussian PBR decomposition as a conditional 3D point-cloud diffusion problem. By fundamentally decoupling geometry from appearance and introducing a multi-scale cross-view conditioning mechanism, our approach rigorously disentangles intrinsic albedo from specular highlights without the need for proxy meshes or continuous UV parameterization. Furthermore, the pipeline replaces per-scene joint optimization of illumination and BRDF with a learned diffusion pass followed by a short target-driven distillation. We believe this paradigm provides a template for the future of 3D generative asset production.


\bibliographystyle{ACM-Reference-Format}
\bibliography{sample-base}

\begin{figure*}[t]
  \centering
  \includegraphics[width=0.9\textwidth] {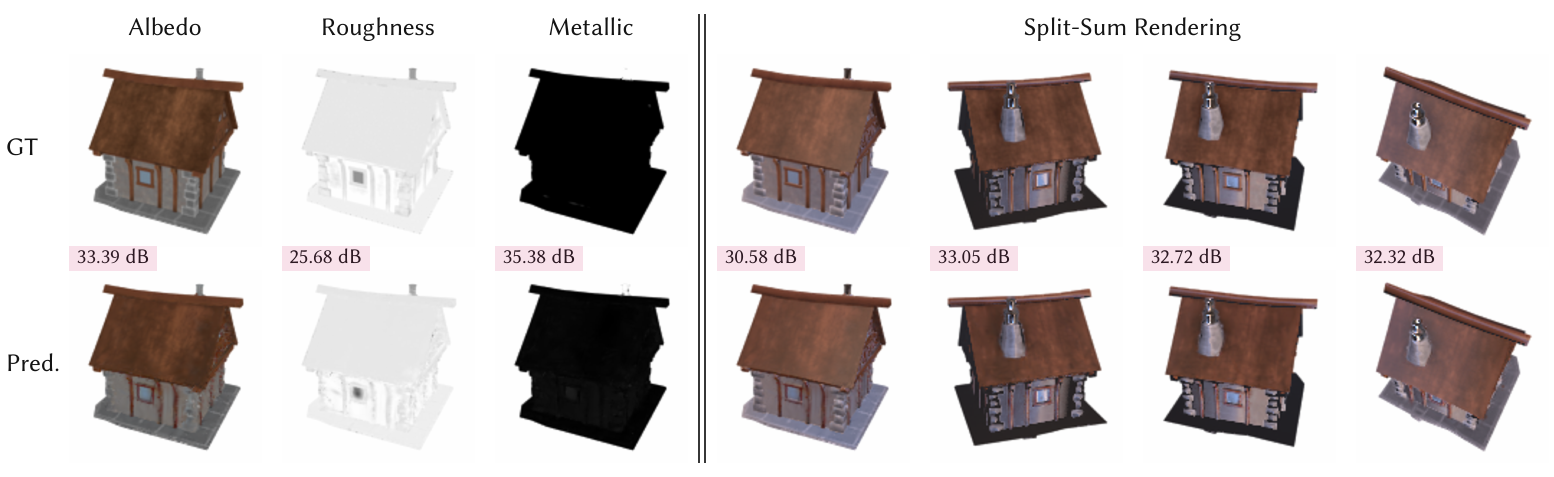}\\[-0.35cm] 
  \includegraphics[width=0.9\textwidth]{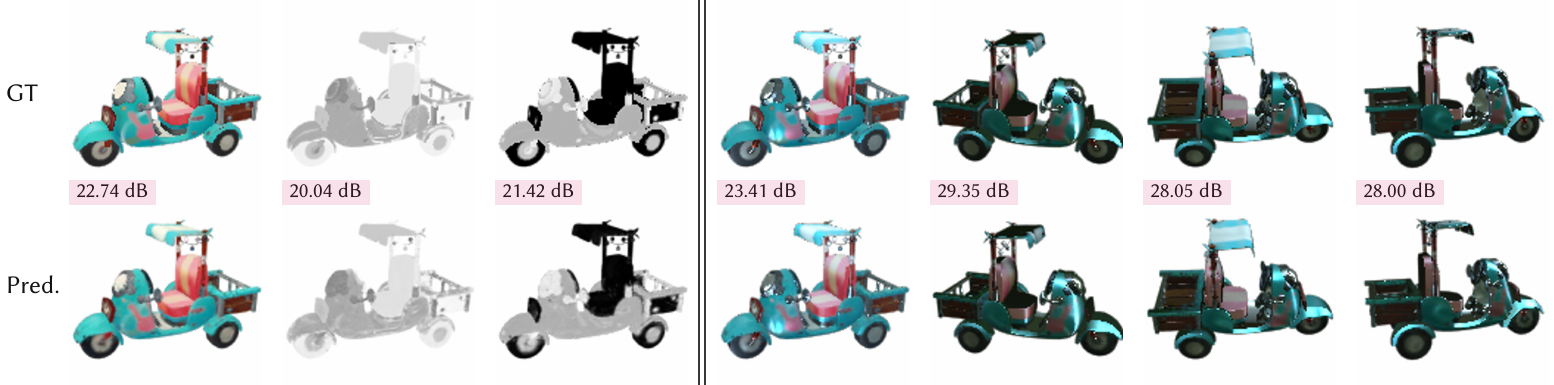}\\[-0.32cm] 
  \includegraphics[width=0.9\textwidth]{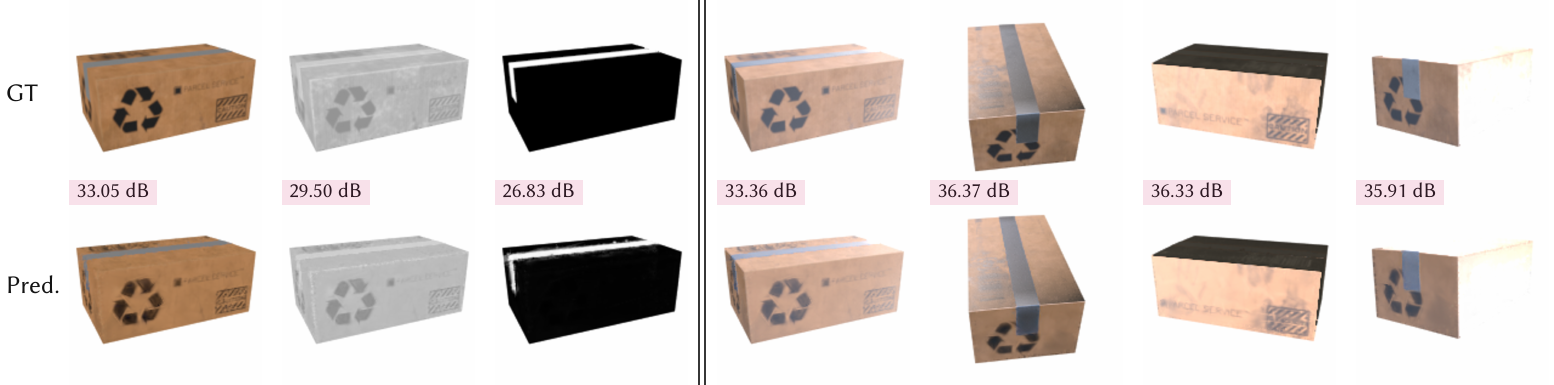}\\[-0.15cm] 
  \includegraphics[width=0.9\textwidth]{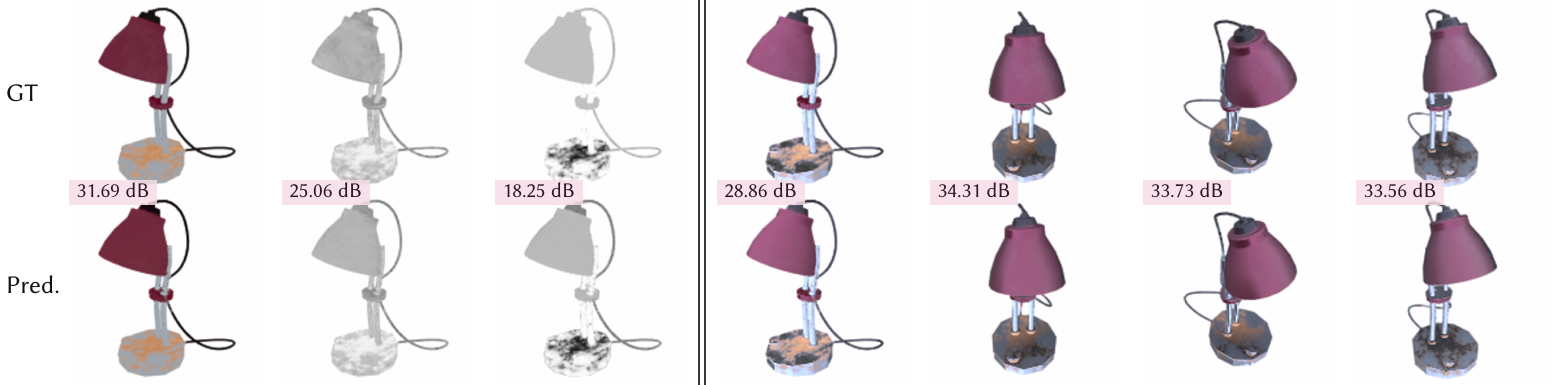}\\[-0.1cm]
  \includegraphics[width=0.9\textwidth]{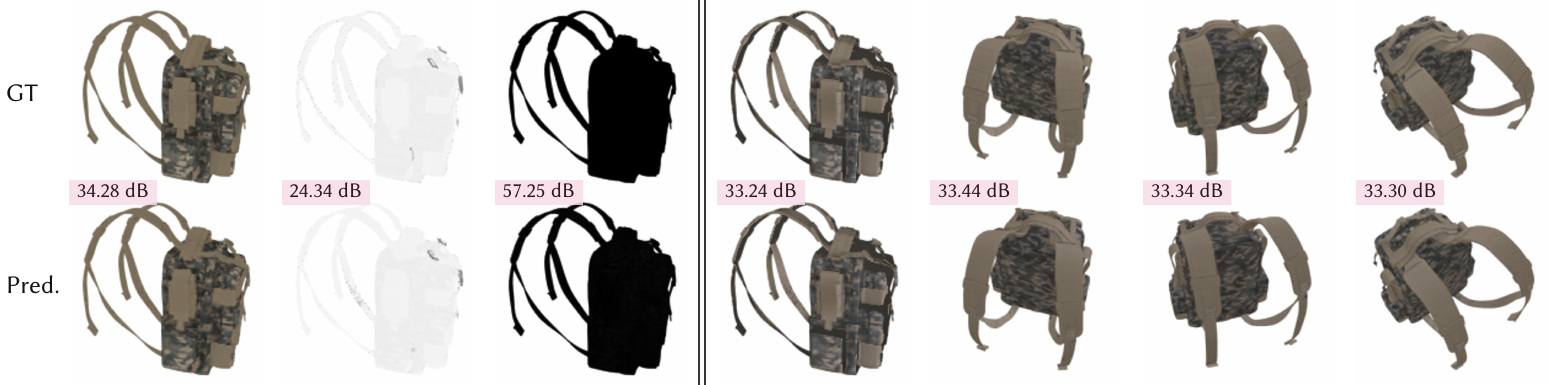}
  \caption{\textbf{Comprehensive Visual Evaluation on Diverse 3D Assets.}  The left panel showcases the disentangled intrinsic maps (Albedo, Roughness, and Metallic), while the right panel demonstrates relighting fidelity via Split-Sum Rendering across four novel environment conditions. The corresponding PSNR metrics (in dB) are reported between the GT and Pred.\ rows for each asset. Across varying material types—ranging from metals and plastics to cardboard and fabrics—our method consistently maintains high structural fidelity and physically accurate specular responses.}
  \label{fig:relight_main}
\end{figure*}

\end{document}